\documentclass{article}
\usepackage{spconf,amsmath,graphicx,hyperref}
\hypersetup{hidelinks}
\usepackage{xcolor}
\definecolor{myblue}{RGB}{0,70,140}
\hypersetup{
    colorlinks=true,
    linkcolor=myblue,
    citecolor=myblue,
    urlcolor=myblue
}
\usepackage{booktabs}
\usepackage{multirow}
\usepackage{amsmath,amsfonts}
\usepackage{algorithm}
\usepackage{array}
\usepackage{textcomp}
\usepackage{stfloats}
\usepackage{url}
\usepackage{verbatim}
\usepackage{graphicx}
\usepackage{cite}
\usepackage{xspace}
\usepackage{subcaption}
\usepackage{colortbl}
\usepackage{tikz}
\usepackage{cuted}
\usepackage{capt-of}

\usepackage[normalem]{ulem}
\usepackage{cleveref}
\crefname{section}{\S}{\S\S}
\Crefname{section}{\S}{\S\S}
\crefname{figure}{fig.}{figs.}
\Crefname{figure}{Fig.}{Figs.}
\crefname{table}{tab.}{tabs.}
\Crefname{table}{Tab.}{Tabs.}
\crefname{equation}{eq.}{eqs.}
\Crefname{equation}{Eq.}{Eqs.}
\crefname{algorithm}{alg.}{algs.}
\Crefname{algorithm}{Alg.}{Algs.}
\makeatletter
\DeclareRobustCommand\onedot{\futurelet\@let@token\@onedot}
\def\@onedot{\ifx\@let@token.\else.\null\fi\xspace}
\newcommand{\parhead}[1]{\noindent\textbf{#1}\xspace}

\makeatother
\definecolor{color1}{HTML}{ECF4F9}
\definecolor{color2}{HTML}{FFF1E0}
\definecolor{color3}{HTML}{ECF4E9}
\definecolor{myzishe}{HTML}{E6E6FA}
\usepackage{enumitem}

\newcommand{\ourmethod}{{\tt \textit{\textbf{S2A}}}\xspace}
\title{\ourmethod: Semantic-to-Spatial Alignment for Alignment-Free RGB-T Salient Object Detection}

\name{Qiangqiang Zhou$^{1}$, Yang Luo$^{1}$*\thanks{* Corresponding author: \{luo\_yang, jiawei\_xu\}@jxnu.edu.cn}
, Yong Chen$^{1}$
, Jiawei Xu$^{1}$*}
\address{$^{1}$Jiangxi Normal University
}
\begin{document}
\maketitle



\begin{abstract}
Alignment-free RGB-T salient object detection (RGB-T SOD) aims to identify salient objects from unregistered RGB and thermal image pairs without costly pre-alignment. However, spatial misalignment breaks pixel-wise correspondence and causes feature contamination during cross-modal fusion. 
To address this issue, we propose \ourmethod, a semantic-to-spatial alignment framework for alignment-free RGB-T SOD. Specifically, a global-guided hierarchical fusion module (GGHF) first exploits global semantic guidance to suppress background interference and refine hierarchical intra-modal features. Subsequently, the alignment-free cross-modal channel attention module (AFCA) globally exchanges complementary semantic information through channel-wise interaction, effectively overcoming the interference caused by local spatial misalignments. Finally, a spatial deformable cross-attention module (SDCA) predicts adaptive sampling offsets to recover local cross-modal spatial correspondence.
Through this semantic-to-spatial paradigm, \ourmethod first enables reliable cross-modal semantic interaction and subsequently performs local spatial calibration, effectively reducing misalignment-induced feature contamination.
Without bells and whistles, \ourmethod achieves highly competitive performance on multiple public alignment-free RGB-T benchmarks, demonstrating its effectiveness in alleviating misalignment-induced feature contamination.
\end{abstract}

\section{Introduction}
RGB-T salient object detection (RGB-T SOD) aims to identify and segment salient objects by jointly exploiting RGB and thermal modalities~\cite{UniSOD,DifferSeg,xu2026hvpnet,SOMANet}. RGB images provide rich appearance information, while thermal images offer illumination-invariant responses. However, RGB-T pairs captured in real-world scenarios often suffer from spatial misalignment due to differences in sensor positions, viewpoints, and imaging characteristics~\cite{DCNet,MVT,VTSalNet}.
To address this issue, recent alignment-free RGB-T SOD methods~\cite{TPS-SCL,SACNet,PTF} have attempted to alleviate geometric discrepancies through feature alignment~\cite{AlignSal,FMTrack}, spatial transformation~\cite{HSMNet,AMNet}, or cross-modal interaction~\cite{AAV-oriented,Tracking}. Although these methods mitigate misalignment to some extent, their cross-modal fusion processes still rely on explicit spatial correspondence, making feature interaction vulnerable to unreliable local matching.
\begin{figure}[t]
    \centering
    \includegraphics[width=\columnwidth]{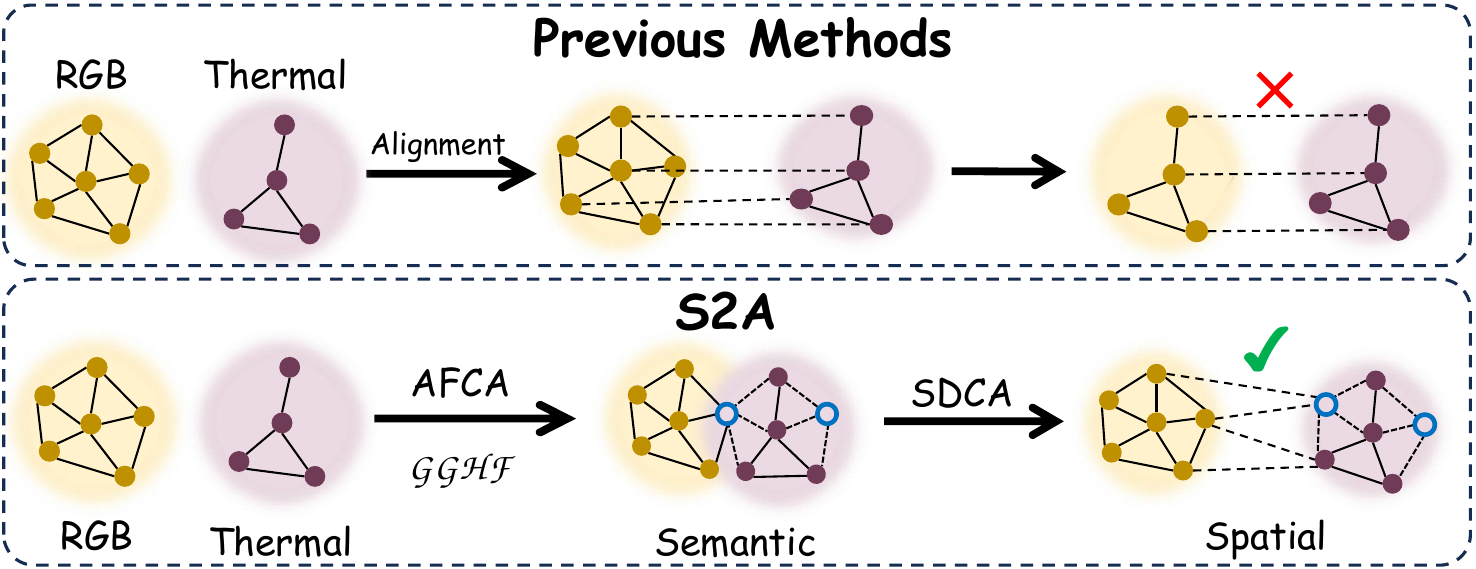}
    \caption{Illustration of semantic-to-spatial alignment. Instead of directly matching misaligned RGB-T features in the spatial domain~\cite{AlignSal,DML,HSMNet}, \ourmethod first refines intra-modal representations with GGHF, establishes cross-modal semantic correspondence through AFCA, and then recovers spatial correspondence via SDCA.}
    \label{fig:image1}
    \vspace{-5mm}
\end{figure}

In unaligned RGB-T pairs, features at the same spatial location may originate from different physical regions, making pixel-level fusion or spatial cross-attention prone to erroneous cross-modal matching. Such mismatches can propagate through the network and degrade multimodal fusion. To address this issue, we propose \ourmethod, a semantic-to-spatial alignment framework for alignment-free RGB-T SOD. As illustrated in \Cref{fig:image1}, \ourmethod avoids premature spatial interaction by first establishing semantic correspondence and then progressively recovering spatial correspondence.

Specifically, we first introduce a global-guided hierarchical fusion module (GGHF) to refine intra-modal representations by using high-level semantics to suppress background interference and selectively integrate multi-level features. Based on these refined representations, an alignment-free cross-modal attention module (AFCA) exchanges complementary semantics via macroscopic channel correlations, effectively bypassing strict pixel-wise alignment. After reliable semantic interaction is established, a spatial deformable cross-attention module (SDCA) further predicts adaptive sampling offsets to recover fine-grained local correspondence between the two modalities. 

\begin{figure*}[t] 
    \centering
    \includegraphics[width=0.8\linewidth]{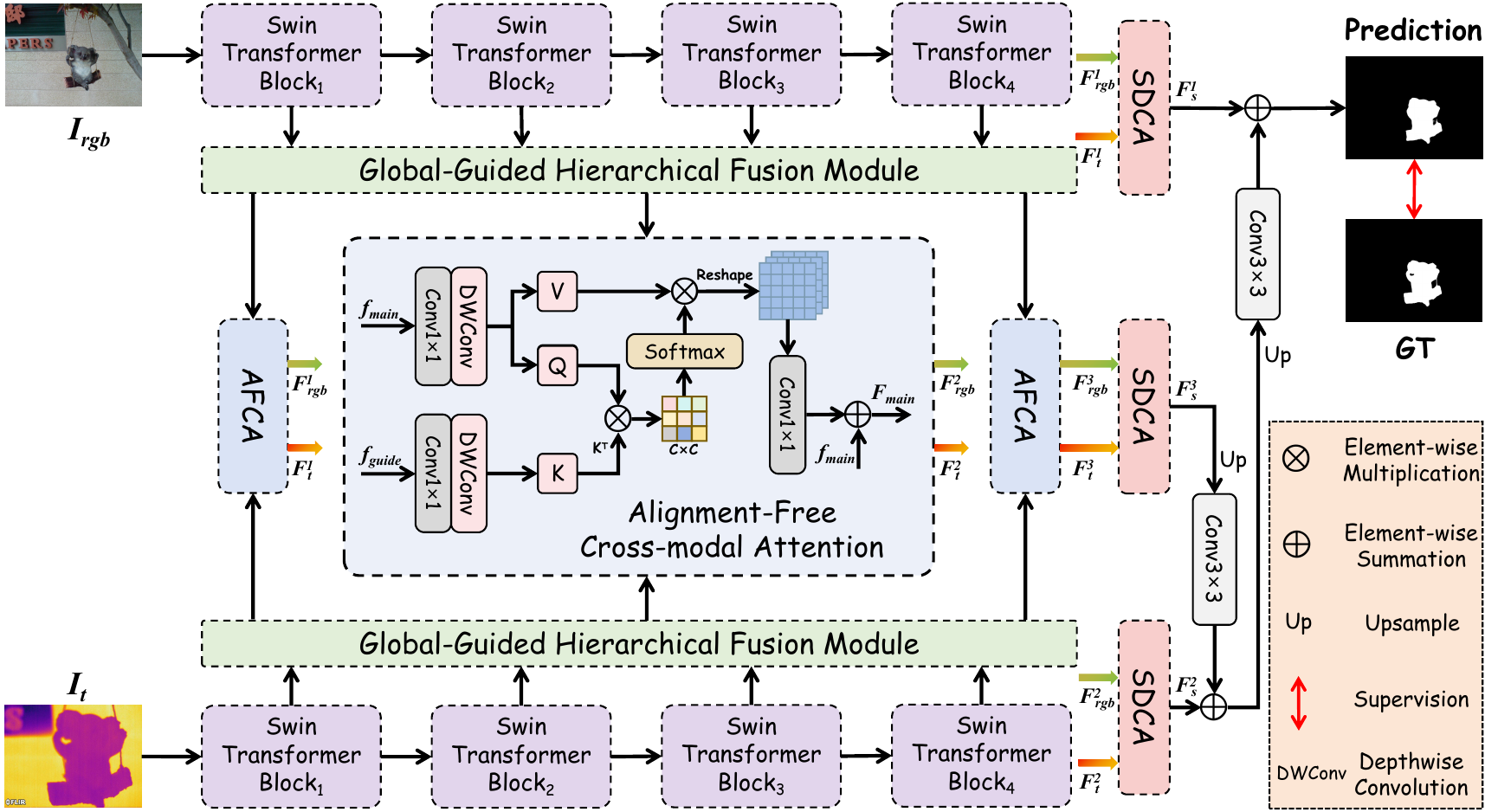}
    \caption{Overall architecture of \ourmethod. The framework employs global-guided hierarchical fusion module (GGHF) for intra-modal feature refinement, alignment-free cross-modal attention module (AFCA) for alignment-free channel-wise cross-modal interaction, and spatial deformable cross-attention module (SDCA) for spatial calibration to produce the prediction map.}
    \label{fig:image2}
    \vspace{-4mm}
\end{figure*}

In summary, our main contributions are as follows:
\begin{itemize}
[leftmargin=*,itemsep=0em,topsep=0em,parsep=0em]
    \item We propose \ourmethod, a semantic-to-spatial alignment framework for alignment-free RGB-T SOD, which establishes cross-modal semantic correspondence before recovering spatial correspondence, reducing the feature contamination.

    \item We develop a global-guided hierarchical fusion module (GGHF) and an alignment-free cross-modal attention module (AFCA) to establish reliable cross-modal semantic correspondence before spatial alignment.

\item We further introduce a spatial deformable cross-attention module (SDCA) to recover local spatial correspondence after semantic interaction, forming a progressive semantic-to-spatial alignment paradigm for robust RGB-T fusion under spatial misalignment.
    
\item Extensive experiments across multiple public alignment-free RGB-T benchmarks verify the effectiveness, robustness, and generalizability of the proposed \ourmethod.
    
\end{itemize}

\section{Method}

\parhead{Overall Architecture.}
As shown in \Cref{fig:image2}, given an unaligned RGB-T image pair $\{I_{rgb}, I_t\}$, \ourmethod employs two parallel encoders to extract hierarchical features from the two modalities. Global-guided hierarchical fusion module (GGHF)  first refines the intra-modal representations under high-level semantic guidance. Alignment-free cross-modal attention module (AFCA) then performs channel-wise cross-modal interaction, effectively alleviating the reliance on strict pixel-wise correspondence. Subsequently, spatial deformable cross-attention module (SDCA) dynamically samples spatial locations to recover local cross-modal alignment. Finally, the refined features are progressively aggregated to produce the final prediction map.


\begin{figure*}[t]
    \centering
    \begin{subfigure}[t]{0.48\textwidth}
        \centering
        \includegraphics[width=\linewidth]{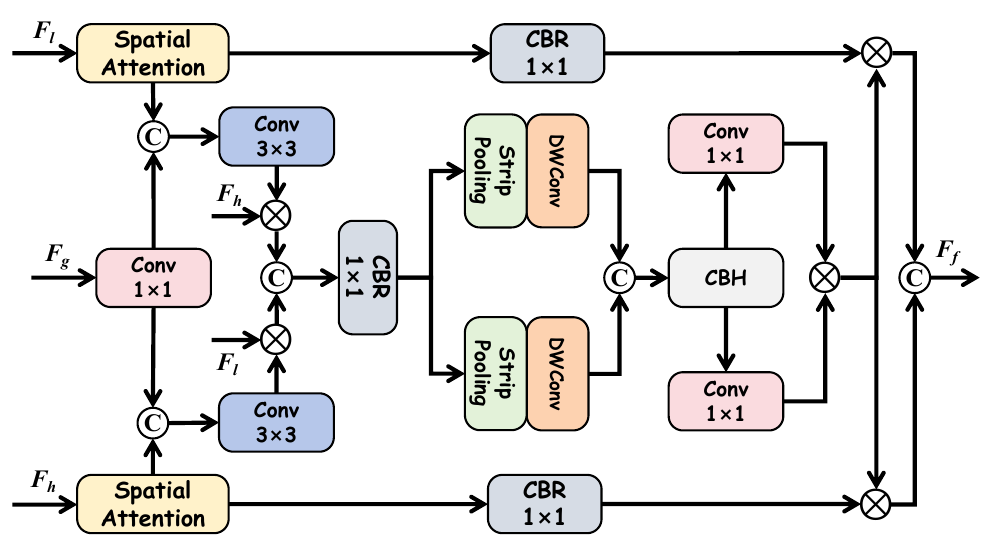}
        \caption{The detailed architecture of the proposed GGHF module.}
        \label{fig:ghfm}
    \end{subfigure}
    \hfill
    \begin{subfigure}[t]{0.48\textwidth}
        \centering
        \includegraphics[width=\linewidth]{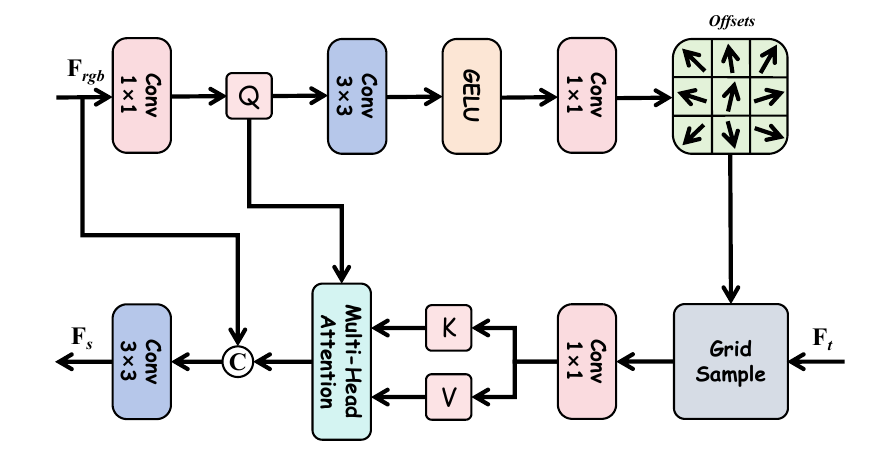}
        \caption{The detailed architecture of the proposed SDCA module.}
        \label{fig:sdcm}
    \end{subfigure}
    \caption{Illustration of the proposed \textbf{G}lobal-\textbf{G}uided \textbf{H}ierarchical \textbf{F}usion and \textbf{S}patial \textbf{D}eformable \textbf{C}ross-\textbf{A}ttention modules.}
    \label{fig:modules}
    \vspace{-3mm}
\end{figure*}
\parhead{Global-Guided Hierarchical Fusion Module (GGHF).}
GGHF refines intra-modal representations by exploiting global semantics to suppress background interference and coordinate hierarchical features.
Given a low-level feature $F_l$, a high-level feature $F_h$, and a global feature $F_g$, we first obtain local and global guidance:
\begin{equation}
M_l=\mathcal{SA}(F_l),\quad
M_h=\mathcal{SA}(F_h),\quad
M_g=\mathcal{C}_{1\times1}(F_g).
\end{equation}

The global guidance is then combined with local cues for bidirectional hierarchical refinement:
\begin{equation}
\hat F_h=F_h\odot \phi([M_l,M_g]),\qquad
\hat F_l=F_l\odot \phi([M_h,M_g]),
\end{equation}
where $\phi(\cdot)$ is a $3\times3$ convolution. We then aggregate the refined features into a shared hierarchical attention:
\begin{equation}
A=
\Psi\left(
\operatorname{CBR}([\hat F_h,\hat F_l])
\right),
\end{equation}
where $\Psi(\cdot)$ consists of strip pooling, depth-wise convolution, CBH, and $1\times1$ convolutions.
Finally, $A$ adaptively enhances the hierarchical representations:
\begin{equation}
F_f=
\left[
\operatorname{CBR}(M_l)\odot A,\,
\operatorname{CBR}(M_h)\odot A
\right],
\end{equation}
where CBR and CBH denote Conv-BN-ReLU and Conv-BN-Hardswish blocks, respectively.

\parhead{Alignment-Free Cross-modal Attention Module (AFCA).}
To enable reliable cross-modal semantic interaction that is highly robust to spatial shifts, we introduce the AFCA module. Specifically, given \(f_{\mathrm{main}}\) and \(f_{\mathrm{guide}}\), AFCA uses \(f_{\mathrm{main}}\) for \(Q,V\) and \(f_{\mathrm{guide}}\) for \(K\):
\begin{equation}
 (Q,V)=
 DWC_{d=2}(C_{1\times1}(f_{main})),
 \end{equation}
 \begin{equation}
K=DWC_{d=2}(C_{1\times1}f_{guide}),
 \end{equation}
 where $DWC_{d=2}$ denotes depth-wise convolution with dilation rate 2.
 After flattening the spatial dimensions, channel-wise attention is computed as
 \begin{equation}
 A=\mathrm{Softmax}\left(\tau QK^{T}\right),
 \end{equation}
 where $\tau$ is a learnable temperature parameter. Since $Q\in\mathbb{R}^{C\times HW}$ and $K^{T}\in\mathbb{R}^{HW\times C}$, the resulting $A\in\mathbb{R}^{C\times C}$ captures macroscopic channel correlations via global spatial aggregation. The output is then obtained as
 \begin{equation}
 F_{\mathrm{main}}
 =
 f_{main}+\mathcal{C}_{1\times1}(AV).
 \end{equation}



\parhead{Spatial Deformable Cross-Attention Module (SDCA).}
To recover local spatial correspondence under misalignment, we introduce SDCA to adaptively sample complementary features for cross-modal spatial calibration.
Let $F_{rgb}$ and $F_t$ denote the features from the RGB and thermal modalities, respectively. We first project $F_{rgb}$ through a $1\times1$ convolution to obtain the query feature $Q$, which is then used to predict spatial sampling offsets:
\begin{equation}
\Delta p =
\operatorname{Scale}\left(
\tanh\left(
\mathcal{C}_{1\times1}
\left(
\delta\left(\mathcal{C}_{3\times3}(Q)\right)
\right)
\right)
\right),
\end{equation}
where $\delta$ denotes GELU activation and $\operatorname{Scale}(\cdot)$ constrains the offset magnitude.
The predicted offsets are added to the base sampling grid to resample the thermal feature:
\begin{equation}
\widetilde{F}_t =
\mathcal{G}\left(
F_t,\,
p_{\mathrm{base}}+\Delta p
\right),
\end{equation}
where $\mathcal{G}(\cdot)$ denotes bilinear grid sampling. 
The spatially calibrated feature $\widetilde{F}_t$ is further projected to generate the key $K$ and value $V$, while $Q$ is directly used as the query for multi-head cross-attention:
\begin{equation}
F_{\mathrm{att}}
=
\operatorname{MHA}(Q,K,V).
\end{equation}
Finally, the attended feature is concatenated with the RGB feature and refined by a $3\times3$ convolution:
\begin{equation}
F_{\mathrm{s}}
=
\mathcal{C}_{3\times3}
\left(
[F_{rgb},F_{\mathrm{att}}]
\right).
\end{equation}

\parhead{Decoder and Loss Function.}
The fused multi-scale features are progressively aggregated through a lightweight top-down decoder~\cite{SACNet} with bilinear upsampling and convolutional refinement to generate the final prediction map. For a fair comparison, following previous methods~\cite{zhou2026tdfnet,zhou2026vico,TPS-SCL,TP-Seg}, we employ the standard binary cross-entropy  loss, smoothness loss, and soft Dice loss as the training objective and optimize the entire network end-to-end.

\section{Experiments}
\begin{table*}[t]
    \centering
    \renewcommand{\arraystretch}{1.1} 
    \setlength{\tabcolsep}{5pt} 
    \caption{Quantitative comparisons on two unaligned and three weakly aligned datasets. The best results are marked in \textbf{bold}.}
    \vspace{-2mm}
    \label{tab:1}
    \resizebox{0.95\textwidth}{!}{ 
    \begin{tabular}{c|c|ccc|ccc|ccc|ccc|ccc}
        \toprule[1.5pt] 
        \multirow{2}{*}{Method} & \multirow{2}{*}{Backbone} & \multicolumn{3}{c|}{UVT20K~\cite{PCNet}} & \multicolumn{3}{c|}{UVT2000~\cite{SACNet}} & \multicolumn{3}{c|}{un-VT5000~\cite{DCNet}} & \multicolumn{3}{c|}{un-VT1000~\cite{DCNet}} & \multicolumn{3}{c}{un-VT821~\cite{DCNet}} \\ \cmidrule{3-17}
        
        & & \cellcolor{color1}$S_m\uparrow$ 
        & \cellcolor{color2}$E_m\uparrow$ 
        & \cellcolor{color3}$F_m\uparrow$ 
        & \cellcolor{color1}$S_m\uparrow$ 
        & \cellcolor{color2}$E_m\uparrow$ 
        & \cellcolor{color3}$F_m\uparrow$ 
        & \cellcolor{color1}$S_m\uparrow$ 
        & \cellcolor{color2}$E_m\uparrow$ 
        & \cellcolor{color3}$F_m\uparrow$ 
        & \cellcolor{color1}$S_m\uparrow$ 
        & \cellcolor{color2}$E_m\uparrow$ 
        & \cellcolor{color3}$F_m\uparrow$ 
        & \cellcolor{color1}$S_m\uparrow$ 
        & \cellcolor{color2}$E_m\uparrow$ 
        & \cellcolor{color3}$F_m\uparrow$ \\ 
        \midrule 
        SwinNet~\cite{SwinNet} & SwinB~\cite{Backbone} & 0.841 & 0.850 & 0.728 & 0.790 & 0.780 & 0.592 & 0.837 & 0.901 & 0.767 & 0.853 & 0.871 & 0.802 & 0.854 & 0.903 & 0.783 \\
        OSRNet~\cite{OSRNet} & VGG16~\cite{VGG16} & 0.807 & 0.835 & 0.730 & 0.741 & 0.764 & 0.567 & 0.800 & 0.846 & 0.741 & 0.871 & 0.885 & 0.833 & 0.814 & 0.848 & 0.734 \\
        TNet~\cite{TNet} & ResNet50~\cite{ResNet50} & 0.856 & 0.876 & 0.783 & 0.792 & 0.782 & 0.610 & 0.858 & 0.905 & 0.807 & 0.868 & 0.880 & 0.827 & 0.876 & 0.913 & 0.829 \\
        DCNet~\cite{DCNet} & VGG16~\cite{VGG16} & 0.808 & 0.853 & 0.776 & 0.767 & 0.808 & 0.632 & 0.812 & 0.879 & 0.803 & 0.858 & 0.880 & 0.850 & 0.817 & 0.869 & 0.793 \\
        HRTrans~\cite{HRTrans} & HRFormer~\cite{HRFormer} & 0.852 & 0.839 & 0.710 & 0.758 & 0.706 & 0.525 & 0.872 & 0.899 & 0.780 & 0.869 & 0.872 & 0.785 & 0.873 & 0.901 & 0.782 \\
        MCFNet~\cite{MCFNet} & ResNet50~\cite{ResNet50} & 0.842 & 0.875 & 0.800 & 0.774 & 0.784 & 0.621 & 0.836 & 0.892 & 0.809 & 0.876 & 0.886 & 0.850 & 0.850 & 0.893 & 0.818 \\
        LSNet~\cite{LSNet} & MobileNet-v2~\cite{MobileNet-v2} & 0.834 & 0.828 & 0.699 & 0.763 & 0.711 & 0.527 & 0.847 & 0.892 & 0.767 & 0.868 & 0.875 & 0.797 & 0.842 & 0.875 & 0.754 \\
        CAVER~\cite{CAVER} & ResNet50~\cite{ResNet50} & 0.861 & 0.876 & 0.790 & 0.786 & 0.782 & 0.616 & 0.850 & 0.893 & 0.805 & 0.873 & 0.881 & 0.838 & 0.818 & 0.856 & 0.780 \\
        LAFB~\cite{LAFB} & Res2Net50~\cite{Res2Net50} & 0.847 & 0.858 & 0.757 & 0.778 & 0.774 & 0.594 & 0.851 & 0.908 & 0.803 & 0.862 & 0.880 & 0.824 & 0.834 & 0.889 & 0.791 \\
        SACNet~\cite{SACNet} & SwinB~\cite{Backbone} & 0.829 & 0.819 & 0.709 & 0.795 & 0.792 & 0.601 & 0.872 & 0.899 & 0.780 & 0.852 & 0.868 & 0.803 & 0.876 & 0.916 & 0.812 \\
        PCNet~\cite{PCNet} & SwinB~\cite{Backbone} & 0.872 & 0.897 & 0.822 & 0.819 & \textbf{0.851} & 0.686 & 0.879 & 0.936 & 0.861 & 0.922 & 0.947 & 0.904 & 0.893 & 0.936 & 0.869 \\
        RA-SOD~\cite{RA-SOD} & Res2Net-50~\cite{Res2Net50} & 0.868 & 0.888 & 0.830 & 0.811 & 0.825 & 0.677 & 0.874 & 0.917 & 0.852 & 0.914 & 0.935 & 0.900 & 0.870 & 0.915 & 0.849 \\
        TPS-SCL~\cite{TPS-SCL} & SwinB~\cite{Backbone} & 0.886 & 0.897 & 0.837 & 0.827 & 0.834 & 0.693 & 0.893 & 0.940 & 0.873 & 0.930 & 0.948 & 0.914 & 0.899 & 0.936 & 0.873 \\ \midrule 
        
        \rowcolor{myzishe}
        \ourmethod & SwinB~\cite{Backbone} & \textbf{0.888} & \textbf{0.902} & \textbf{0.844} & \textbf{0.835} & 0.837 & \textbf{0.702} & \textbf{0.901} & \textbf{0.942} & \textbf{0.882} & \textbf{0.937} & \textbf{0.950} & \textbf{0.917} & \textbf{0.911} & \textbf{0.937} & \textbf{0.876} \\ 
        \bottomrule[1.5pt] 
    \end{tabular}
    }
\vspace{-4mm}
\end{table*}

\begin{table}[t]
  \centering
  \caption{Ablation studies of the proposed modules.}
  \vspace{-3mm}
  \label{tab:2}
  \renewcommand{\arraystretch}{1.2} 
  \setlength{\tabcolsep}{0.8mm} 
  \resizebox{0.9\linewidth}{!}{ 
  \begin{tabular}{c|ccc|ccc|ccc}
    \toprule[1.5pt]
     \multirow{2}{*}[-0.5ex]{Method} & \multicolumn{3}{c|}{UVT20K~\cite{PCNet}} & \multicolumn{3}{c|}{UVT2000~\cite{SACNet}} & \multicolumn{3}{c}{un-VT5000~\cite{DCNet}} \\
    \cmidrule(lr){2-4} \cmidrule(lr){5-7} \cmidrule(lr){8-10}
     & \cellcolor{color1}$S_m\uparrow$ 
     & \cellcolor{color2}$E_m\uparrow$ 
     & \cellcolor{color3}$F_m\uparrow$ 
     & \cellcolor{color1}$S_m\uparrow$ 
     & \cellcolor{color2}$E_m\uparrow$ 
     & \cellcolor{color3}$F_m\uparrow$ 
     & \cellcolor{color1}$S_m\uparrow$ 
     & \cellcolor{color2}$E_m\uparrow$ 
     & \cellcolor{color3}$F_m\uparrow$ \\
    \midrule
     w/o GGHF & 0.865 & 0.883 & 0.788 & 0.810 & 0.817 & 0.650 & 0.882 & 0.930 & 0.840 \\
     w/o AFCA & 0.866 & 0.874 & 0.776 & 0.803 & 0.809 & 0.643 & 0.873 & 0.932 & 0.842 \\
     w/o SDCA & 0.861 & 0.843 & 0.735 & 0.803 & 0.799 & 0.616 & 0.869 & 0.925 & 0.826 \\
     Spatial to Semantic & 0.885 & 0.871 & 0.788 & 0.830 & 0.821 & 0.672 & 0.894 & 0.933 & 0.869 \\
    \midrule
    \rowcolor{myzishe}
     \ourmethod & \textbf{0.888} & \textbf{0.902} & \textbf{0.844} & \textbf{0.835} & \textbf{0.837} & \textbf{0.702} & \textbf{0.901} & \textbf{0.942} & \textbf{0.882} \\
    \bottomrule[1.5pt]
  \end{tabular}
  } 
  \vspace{-4mm}
\end{table}

\subsection{Evaluation}
\parhead{Datasets and Metrics.}
For a comprehensive evaluation of \ourmethod, we adopt the same training set~\cite{PCNet} as previous methods~\cite{PCNet,TPS-SCL} and conduct testing on both unaligned~\cite{PCNet,SACNet} and weakly aligned~\cite{DCNet} datasets. We employ three widely used evaluation metrics to assess model performance, including structure-measure \colorbox{color1}{$S_m$}~\cite{Sm}, enhanced-alignment measure \colorbox{color2}{$E_m$}~\cite{EM}, and F-measure \colorbox{color3}{$F_m$}~\cite{Fmeasure}.

\parhead{Implementation Details.}
\ourmethod is implemented on a single NVIDIA RTX 4090 GPU with Swin-B~\cite{Backbone} as the backbone. We use AdamW~\cite{adamW} with a learning rate of $5\times10^{-5}$ and a batch size of 2 for 200 epochs. All images are resized to $384\times384$ during training and inference.

\begin{figure}[t]
    \centering
    \includegraphics[width=0.9\linewidth]{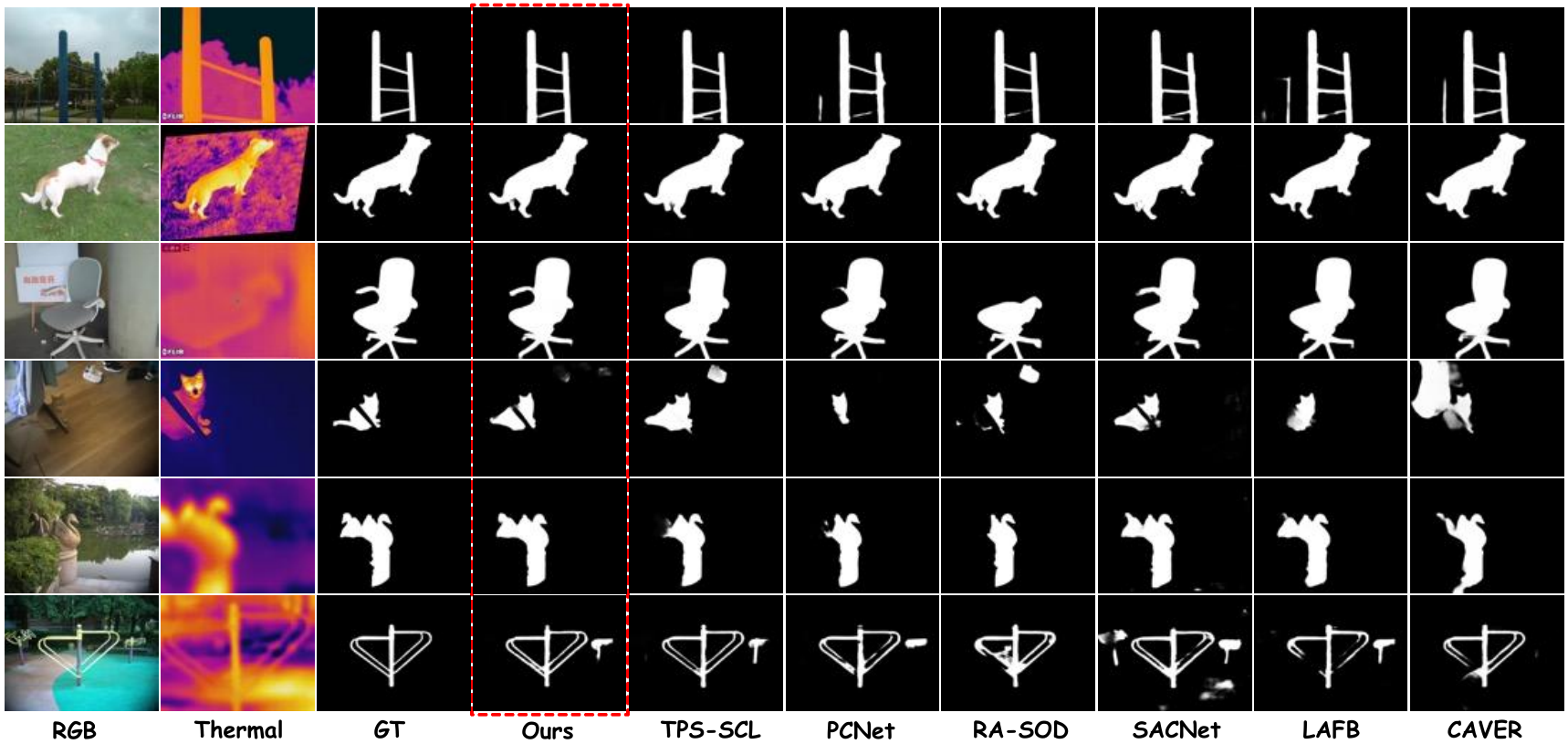}
    \vspace{-2mm}
    \caption{Visual comparisons with other SOTA methods.}
    \label{fig:visual_comparison}
    \vspace{-5mm}
\end{figure}

\parhead{Comparison with State-of-the-arts.}
\Cref{tab:1} presents the quantitative comparison between \ourmethod and current SOTA methods on two unaligned and three weakly aligned RGB-T SOD datasets. Overall, \ourmethod achieves the best performance on almost all evaluation metrics, demonstrating strong generalization under different degrees of spatial misalignment. Specifically, compared with the recent state-of-the-art method TPS-SCL~\cite{TPS-SCL}, \ourmethod achieves comprehensively better results across all three metrics on all five evaluated datasets. On the UVT2000 dataset, although the $E_m$ score of \ourmethod is second to PCNet~\cite{PCNet}, it secures the highest $S_m$ and $F_m$ scores, maintaining its overall superiority. The qualitative results in \Cref{fig:visual_comparison} further demonstrate the effectiveness of \ourmethod under challenging conditions, including cluttered backgrounds, weak alignment, low illumination, degraded thermal signals, thermal diffusion, and fine structures. \ourmethod consistently produces more accurate saliency maps with more complete object structures and sharper boundaries, further demonstrating its robustness to cross-modal spatial misalignment.

\vspace{-2mm}
\subsection{Ablation Studies}
\ourmethod progressively alleviates cross-modal spatial misalignment through the complementary effects of \textit{intra-modal feature refinement}, \textit{cross-modal semantic interaction}, and \textit{spatial correspondence recovery}. As shown in \Cref{tab:2}, 
to verify the effectiveness of \uline{intra-modal feature refinement},we replace the global-guided hierarchical fusion module (GGHF) with standard convolutional blocks, which leads to noticeably increased false-positive responses in cluttered backgrounds. This demonstrates the importance of global semantic guidance for suppressing background interference and refining intra-modal representations. Next, replacing alignment-free cross-modal attention module (AFCA) with simple channel-wise concatenation causes a clear performance drop, confirming the critical role of AFCA in establishing reliable \uline{cross-modal semantic interaction}. Finally, replacing spatial deformable cross-attention module (SDCA) with rigid co-located fusion degrades the performance across all metrics, further validating the necessity of SDCA for \uline{spatial correspondence recovery}. To further validate the semantic-to-spatial paradigm, we reverse the alignment order into spatial-to-semantic. The performance degradation confirms the effectiveness of establishing semantic correspondence before spatial correspondence recovery.

\section{Conclusion}
\label{sec:conclusion}
In this paper, we present \ourmethod for alignment-free RGB-T salient object detection (RGB-T SOD). Instead of directly establishing unreliable pixel-wise correspondence, \ourmethod follows a progressive semantic-to-spatial alignment paradigm, where global-guided hierarchical fusion module, alignment-free cross-modal attention module, and spatial deformable cross-attention module sequentially perform intra-modal representation refinement, cross-modal semantic interaction, and local spatial correspondence recovery. Through this design, \ourmethod effectively alleviates misalignment-induced feature contamination while fully exploiting complementary multimodal information. Extensive experiments on multiple public benchmarks demonstrate the effectiveness of \ourmethod under varying degrees of spatial misalignment. Overall, our results suggest that semantic-to-spatial alignment provides an effective strategy for robust multimodal perception under imperfect spatial correspondence, and we hope \ourmethod can further promote the development of alignment-free RGB-T SOD.

\footnotesize
\bibliographystyle{IEEEbib}
\bibliography{refs}

\end{document}